%% file: main_arxiv.tex
\documentclass{article}

\usepackage{PRIMEarxiv}

\usepackage[utf8]{inputenc} % allow utf-8 input
\usepackage{url}            % simple URL typesetting
\usepackage{booktabs}       % professional-quality tables
\usepackage{graphicx}       % graphics
\usepackage{threeparttable}
\usepackage{comment}
\usepackage{multirow}
\usepackage{enumitem}
\usepackage{tabularx}
\usepackage{subcaption}

\title{MINT - Mainstream and Independent News Text Corpus}

\author{
  Danielle Caled, Paula Carvalho, Mário J. Silva \\
  INESC-ID, Instituto Superior Técnico, Universidade de Lisboa, Portugal \\
  \texttt{\{dcaled,pcc,mjs\}@inesc-id.pt} \\
}

\begin{document}
\maketitle

\begin{abstract}
\input{content/abstract}
\end{abstract}

% keywords can be removed

\input{content/text_propor}

%\section*{Acknowledgments}
%This was was supported in part by......

%Bibliography
\bibliographystyle{unsrt}  
\bibliography{main_arxiv}

\end{document}

%% file: content/abstract.tex
Most corpora approach misinformation as a binary problem, classifying texts as \textit{real} or \textit{fake}. However, they fail to consider the diversity of existing textual genres and types, which present different properties usually associated with credibility. To address this problem, we created MINT, a comprehensive corpus of news articles collected from mainstream and independent Portuguese media sources, over a full year period. MINT includes five categories of content: \textit{hard news}, \textit{opinion} articles, \textit{soft news}, \textit{satirical} news, and \textit{conspiracy} theories. This paper presents a set of linguistic metrics for characterization of the articles in each category, based on the analysis of an annotation initiative performed by online readers. The results show that (i) conspiracy theories and opinion articles present similar levels of subjectivity, and make use of fallacious arguments; (ii) irony and sarcasm are not only prevalent in satirical news, but also in conspiracy and opinion news articles; and (iii) hard news differ from soft news by resorting to more sources of information, and presenting a higher degree of objectivity.
%We truly believe that the release of the MINT corpus will support future exploratory news studies on misinformation detection, and provide new insights to correctly approach this research task.

%% file: content/text_propor.tex
\section{Introduction} 

The detection of misinformation has been increasingly discussed by the Natural Language Processing (NLP) community, in particularly those concerned with the development of linguistic resources, and methods for identifying false or misleading information, generically known as \textit{fake news}. 

Fake news detection focuses predominantly on distinguishing real from fake content \cite{nakamura2020fakeddit}, approaching this issue as a dichotomous problem.
As most of the resources conceived within the scope of misinformation studies comprise only these two categories, they fail to consider the diversity of existing textual genres and types, including soft news and fictional news stories, created for entertaining purposes. In turn, most misinformation corpora consider only the most extreme cases in the credibility spectrum (e.g., hard news collected from mainstream newspapers \textit{vs.} news previously labeled as fake by fact-checking agencies), making the automatic classification task deceptively simple and misaligned with reality. However, credibility should be regarded as a complex construct, presenting several dimensions in a continuum. %Likewise, the attempt to include news articles in only two categories is limiting, given the huge variety of news types and genres, and their communication purposes.
%To correctly handle this problem, it is crucial to first investigate which are the main characteristics underlying different types and genres of content typically classified as real or fake.

This paper presents the MINT (Mainstream and Independent News Text) corpus, which was specifically developed to address the gaps on misinformation corpora, especially for Portuguese. MINT is composed of more than 20 thousand articles collected from 33 Portuguese mainstream and independent media over a whole year, covering different styles, subjects, and serving different communication purposes.

The collected articles are labeled under five categories, namely \textit{hard news}, \textit{opinion}, \textit{soft news}, \textit{satire}, and \textit{conspiracy}. Although far from being exhaustive, this list includes categories presenting different properties that must be taken into account in misinformation studies. For example, hard news stories are supposed to involve a neutral and objective reporting, while opinions are characterized by their inherent subjectivity, which is a relevant feature for distinguishing reliable from unreliable news \cite{holt2019key,zhang2018structured}. On the other hand, soft news usually approach light topics, including sensational, disruptive and entertainment-oriented news, which generally resort to clickbait strategies to attract the readers' attention \cite{perezrosas2018automatic}. Some of these characteristics may also be found in non-credible news articles, namely in satirical news, created for humorous purposes, and conspiracy theories, fabricated to deceive the reader \cite{carvalho2020situational}. 

We discuss the main linguistic properties of each collection, and present the preliminary results of an evaluation on crowdsourced annotations by online news readers of Portuguese media. Those annotations, addressing aspects previously associated with news credibility \cite{carvalho2021what}, can help to understand the main differences among the articles published by the media sources, allowing the development of computational models to correctly approach misinformation detection. 

% In addition, previous works report that readers have some difficulty in differentiating facts from opinions \cite{}, besides noticing some similarity between opinion pieces and texts related to conspiracy theories \cite{}. 

%This paper is organized as follows: We start with a brief related work (Section \ref{sec:related_work}), and then explain how we collected the news articles to assemble the MINT corpus (Section \ref{sec:news_collection}). Section \ref{sec:corpus_organization} addresses the corpus organization, including the harvested news articles, and the annotations collected through a crowdsourcing process. Section \ref{sec:corpus_characterization} presents a set of metrics that will help to perform a linguistic characterization of the classes considered in the corpus. In addition, we present a topic analysis study, which helps to find out which are the main topics (or themes) approached (Section \ref{sec:topic_analysis}). Finally, we discuss the potential use cases of MINT corpus in Section \ref{sec:potential_uses_cases}, pointing out different tasks and applications. The main conclusions are presented in Section \ref{sec:conclusion}. 
%%In this section, we present the methodology used to create this corpus and the corresponding data structure. 

\vspace{-1mm}%Put here to reduce too much white space

\section{Related Work}
\label{sec:related_work}

Binary classification of misinformation has several conceptual problems, including the establishment of a proper definition of real (or credible) news. For example, some authors associate credibility with a greater degree of factuality and a lesser degree of sentiment in the text \cite{fuhr2018information,aker2019corpus,spradling2021protection}; others highlight a variety of aspects, including adherence to journalistic practices/editorial norms, impartial reporting, and the inclusion of statistical data, credible sources, quotes and attributions in text \cite{shoemaker2017news,molina2021fake}.
However, the non-adherence to these standards should not be used by itself as a proxy to infer the text credibility. Accordingly, opinion pieces are expected to be subjective, and to present an emotionally charged tone \cite{golbeck2018fake,molina2021fake,carvalho2021what}. A more paradigmatic example involves satire, which, despite being fictional, mimics the tone, style and appearance of factual news, leading some authors to label this type of content as fake \cite{rubin2015deception}. Other efforts, in turn, argue that satire does not intend to deceive readers and, therefore, should be considered as an independent class when classifying misinformation \cite{horne2017this,potthast2018stylometric}.
% , but they do not intend to deceive readers

To address the unrepresented news categories in misinformation classification, Molina et al. organized a taxonomy differentiating \textit{real news} from a variety of controversial news, namely \textit{hoaxes}, \textit{polarized content}, \textit{satire}, \textit{misreporting}, \textit{opinion}, \textit{persuasive information}, and \textit{citizen journalism} \cite{molina2021fake}.
% Our work is in line with this taxonomy in the sense that MINT's categories contemplate four of the news categories addressed by Molina et al.
Similarly, MINT's news articles are labelled to different categories, among which \textit{hard news, opinion, satire}, and \textit{conspiracy}, corresponding to Molina et al.'s \textit{real news}, \textit{opinion}, \textit{satire}, and \textit{hoaxes}, respectively.
In our study, we have extended the MINT collection with a new nuance of real news, the \textit{soft news} category, representing ``light'' or ``spicy'' stories with a ``low level of substantive informational value'' \cite{lehmanwilzig2010hard}.

Despite the variety of corpora supporting misinformation classification, few linguistic resources are available for Portuguese (e.g., \cite{monteiro2018contributions,moura2021automated}). The Fake.Br corpus includes news articles in just two categories (\textit{true} or \textit{false}) \cite{monteiro2018contributions}. This corpus comprises articles from four fake news sources, and three major news agencies. However, the Fake.Br corpus is strongly biased regarding text length, typos and sentiment \cite{silva2020towards}, which makes the analysis simplistic and the classification less challenging. Moura et al. developed another news articles collection focused on Portuguese \cite{moura2021automated}. This corpus is also a binary resource, with all the genuine articles in this collection scraped from a single source. Therefore, one cannot rely on this collection for misinformation classification due to the lack of representativeness of credible sources. The MINT corpus differs from these two resources by offering greater diversity of news categories and information sources, containing texts from 33 different mainstream and independent media channels. 

The corpora most similar to MINT are the
ones comprising news articles from different topics. The NELA-GT series include articles harvested from different mainstream, hyperpartisan, and conspiracy sources \cite{norregaard2019nela}. FacebookHoax comprises information extracted from Facebook pages, scientific news and conspiracy websites \cite{tacchini2017some}. Like MINT, both NELA-GT and FacebookHoax assign the credibility label based on the source-level reliability. 
Hardalov et al. also assembled a collection consisting of credible, fictitious, and funny news, resorting to similar strategies for building a corpus in a under-resourced language \cite{hardalov2016search}. 
This work is related to ours since it contains news collected from mainstream, satirical and fictional sources, covering different topics, such as politics and lifestyle.

\vspace{-2mm}%Put here to reduce too much white space

\section{Corpus organization}
\label{sec:corpus_organization}

The MINT corpus consists of two different, but complementary, resources, namely, \textit{MINT-articles}, and \textit{MINT-annotations}. 
The main resource, \textit{MINT-articles} corresponds to the entire collection of news articles extracted from mainstream and independent channels. We also provide \textit{MINT-annotations} as a supplementary resource, containing the manual annotations for a subset of the \textit{MINT-articles} collection (Subsection \ref{sec:mint-annotations}), obtained through a crowdsourcing process. With the insights gained from the annotations, we can therefore understand the specific and shared characteristics of the categories included in the corpus, available for the research community at \url{https://github.com/**hidden-for-blind-review**}.

\subsection{MINT-articles}
\label{sec:mint-articles}

The MINT corpus includes 20,278 articles, published from June 1st, 2020 to March 31st, 2021, representing a full year sample of online content published by the Portuguese mainstream and independent media. 
All articles in the MINT corpus were semi-automatically harvested and assigned to a category through the heuristic rules defined bellow. Table \ref{tab:dataset_stats_headlines} presents examples of article headlines from each MINT's category.

\begin{description}[style=unboxed,leftmargin=0cm]
\item[Hard News (6000 documents):] News collected from the \textit{politics}, \textit{society}, \textit{business}, \textit{technology}, \textit{culture}, and \textit{sports} sections from nine mainstream news websites. Since this content is published by reputable news sources, verified by the Portuguese regulatory authority for social communication (ERC\footnote{\url{https://www.erc.pt/pt/listagem-registos-na-erc}}), they were blindly labeled as hard news.

\item[Opinion (6000 documents):] Articles collected from the \textit{opinion} section of 10 mainstream and independent newspapers and magazines. In general, the collected articles discuss controversial and contemporary topics related to events with great notoriety in the mainstream media. 

\item[Soft News (6000 documents):] This category comprises soft news extracted from \textit{celebrity}, \textit{fashion}, \textit{beauty}, \textit{family}, and \textit{lifestyle} sections of six magazines, tabloids and newspaper supplements.

\item[Satire (1029 documents):] Articles extracted from two well-known websites, self-declared as fictional, humorous, and/or satirical in their editorial guidelines. They parody the tone and format of traditional news stories, by exploring the use of rhetorical devices, such as irony and sarcasm.

\item[Conspiracy (1249 documents):] For identifying conspiracy stories, we explored websites that had previously published at least five articles supporting conspiracy theories, particularly about the origin, scale, prevention, diagnosis, and treatment of the COVID-19 pandemic. We resorted to the COVID-19 theme as it is recurring issue, addressed both by the mainstream and independent media during the MINT collection period.
Thus, we investigated the five conspiracy topics regarding the COVID-19 pandemic previously described by Shahsavari et al. \cite{shahsavari2020conspiracy}, and manually inspected a set of candidate websites; only six websites met the selection criteria. The topics covered by these sources are diverse, ranging from \textit{politics}, \textit{economics}, \textit{conflicts}, \textit{health issues}, to \textit{technology}.
\end{description}

%The imbalance among the classes in MINT corpus highlights the difficulty in collecting specific textual genres from the Portuguese media, namely satirical and conspiracy content. Particularly regarding satires, there are few known sources with regular and abundant production of such type of content, which could be due to the low profitability, fragile public engagement, or concerns that satire could be perceived as misinformation \cite{golbeck2018fake}. On the other hand, the difficulty in obtaining conspiracy theory texts is due to the volatile permanence of these domains on the web (short life-cycle), non-indexed conspiracy pages, which are rarely retrieved by search engines, and the use of other digital platforms (e.g., YouTube and Whatsapp) to disseminate conspiracy narratives \cite{soares2021research,caled2021digital}.

\begin{table}[t]
\centering
\caption[]{Illustrative examples of headlines included in MINT.}
\begin{tabular}{p{0.9cm}| p{1.8cm} | p{9cm}}
\toprule
Alias & Category & Examples \\
\midrule
H-N & Hard News & \textit{O que já se sabe sobre a origem da Covid-19?} \\
\hline
OPI & Opinion & 
\textit{Os políticos no palco da pandemia} \\
\hline
S-N & Soft News & \textit{Príncipe Harry surpreende ao aparecer na televisão britânica} \\
\hline
SAT & Satire & \textit{Primavera foi barrada na fronteira e já não chega amanhã} \\
\hline
CON & Conspiracy & \textit{Máscaras faciais representam riscos graves para a saúde} \\
\bottomrule
\end{tabular}
\label{tab:dataset_stats_headlines}
\vspace{-3mm}%Put here to reduce too much white space after your table 
\end{table}

\subsection{MINT-annotations}
\label{sec:mint-annotations}

In order to understand the readers' ability to distinguish the news articles belonging to different categories, we conducted a human assessment study focused on information content indicators \cite{zhang2018structured}. These indicators are commonly used as proxies for assessing news articles credibility, addressing semantic and discourse dimensions, such as the headline accuracy, the presence of reasoning errors, and sentiment intensity \cite{carvalho2021assessing}.
%However, we went beyond the trustworthiness dimension of texts, extending the assessments to MINT's news categories. Also, we resorted to crowdsourced assessment as we wanted to understand how online readers experience distinct textual contents.

The survey was disseminated to the Portuguese community through different news outlets, inviting online readers to assess a news article randomly selected from \textit{MINT-articles}. Together, these annotations compose the  \textit{MINT-annotations}, which includes 750 judgments on 335 different news articles distributed, by category, as follows: 71 hard news, 63 opinion pieces, 66 soft news, 69 satires, and 65 conspiracy articles.
Each assessment was carried out by a different reader, following annotation guidelines \cite{carvalho2021assessing}. The annotators answered two types of questions: 
 
\begin{description}[style=unboxed,leftmargin=0cm]
    \item[i) Dichotomous questions.] Yes/No questions aimed at assessing the presence or absence in text of specific properties, namely the ones related to information sources, subjectivity, irony or sarcasm, and the particular use of strategies addressing personal attack or appeal to fear. 
    \item[ii) Five-point Likert scale questions.] Questions assessing the overall article credibility, and other dimensions on the news headline and body (e.g., the degree of headlines' accuracy, \textit{clickbaitiness}, sentiment intensity, reliability of the sources of information mentioned in text, linguistic accuracy, and sarcasm or irony).
 \end{description}
 
The information provided by online readers can be used to estimate the main similarities and divergences among the various categories of articles included in the MINT corpus, and understand which features are perceived as the most relevant by readers for assessing news credibility \cite{carvalho2021what}.

\vspace{-2mm}%Put here to reduce too much white space

\section{Corpus characterization}
\label{sec:corpus_characterization}

In this section, we present statistics derived from a set of metrics often used in computational linguistics to characterize the MINT news texts (Subsection \ref{sec:linguistic_characterization}). We also go through some insights obtained from the crowdsourced annotations in Subsection \ref{sec:human_annotation}.

\subsection{Linguistic characterization}
\label{sec:linguistic_characterization}

Table \ref{tab:data_stats_style} presents quantitative metrics related to style and text complexity, which estimate the average number of sentences ($\#s$) and words ($\#w$) comprised in the headline and body text. We have also calculated the average number of words per sentence ($w/s$), which may help distinguishing elementary from complex sentences.
We notice that headlines from opinion articles tend to be shorter, while satire headlines are longer, when compared to the headlines belonging to the remaining categories. 
Despite the wide diversity on the body length of articles in each category, the statistics obtained show that satirical news stories are usually short, comprising a restricted number of simple sentences. This may indicate that the story introduced in the headline is not deeply developed in the body text.
In contrast, the most extensive articles are from the conspiracy category, on average, up to three times longer than the articles reporting hard news. This apparently contradicts the previous studies focused on Portuguese stating that false articles are usually shorter than credible articles \cite{monteiro2018contributions,moura2021automated}.
When comparing hard with soft news, we can observe that the former tend to be longer, and use more complex linguistic structures.

\begin{table}[t]
\centering
\caption{Style and complexity characterization of MINT ($\#s$, $\#w$, $w/s$ are the number of sentences, number of words, number of words per sentence, resp.).}
\begin{tabular*}{\textwidth}{l | @{\extracolsep{\fill}} rrrrr|rrrrr}
\toprule
& \multicolumn{5}{c|}{Headline} & \multicolumn{5}{c}{Body Text} \\
\hline
& H-N & OPI & S-N & SAT & CON & H-N & OPI & S-N & SAT & CON \\
%& & avg & std & avg & std & avg & std & avg & std & avg & std \\ 
\midrule
Avg \#s  & 1.12 & 1.08 & 1.12 & 1.01 & 1.03 & 13.72 & 28.31 & 15.61 & 5.29 & 55.49\\ 
Avg \#w & 11.76 & 7.17 & 12.1 & 14.58 & 10.36 & 414.75 & 672.24 & 297.13 & 115.42 & 1372.34 \\
Avg w/s & 10.94 & 6.76 & 11.28 & 14.48 & 10.14 & 32.63 & 25.53 & 20.92 & 27.09 & 26.05 \\
\bottomrule
\end{tabular*}
\label{tab:data_stats_style}
\vspace{-3mm}%Put here to reduce too much white space after your table 
\end{table}

Table \ref{tab:data_stats_linguistic} provides a set of metrics that have been explored in the research on news credibility \cite{zhou2004comparison,zhou2020survey}. 
To generate those statistics, texts were tagged with PoS\footnote{PoS tagging was performed using spaCy API: \url{https://spacy.io/}}, and the sentiment information was estimated using SentiLex \cite{silva2012building}. With regard to sentiment, we only present the information on the headline, since this information did not seem relevant in the characterization of the news body. 

The results indicate that adjectives are less used in sentences from shorter news texts, namely those belonging to soft news and satire categories, while adverbs are chiefly frequent in satirical news. Globally, these modifiers are mostly used in texts where a higher degree of subjectivity is expected, namely in opinion articles and conspiracy theories. Conversely, the hard news, which should be objective and neutral by principle, use comparatively less personal pronouns (only found in quotations or citations included in the news body), and more numerals, which is critical for attesting the text credibility \cite{koetsenruijter2011using}. Conjunctions and punctuation marks (pausality) are also more recurrent in hard news, corroborating the perception of textual cohesion and the idea that authors opt for more complex linguistic constructions. Additionally, sentiment terms are more frequent in headlines from soft news and conspiracies, which are often sensationalist, and employ a emotionally charged tone. On the other hand, soft news use comparatively fewer modal verbs and indefinite pronouns, which support the idea that they adopt a direct and focused narrative. Finally, the data shown in Table \ref{tab:data_stats_linguistic} also suggests that opinion and conspiracy articles are quite similar, with the exception of a slightly more pronounced use of indefinite pronouns in opinion articles.

\begin{table*}[t]
\begin{threeparttable}[]
\centering
\caption{Linguistic characterization of MINT corpus.}
\begin{tabular}{l|ccccccc}
\toprule
Average metrics & H-N & OPI & S-N & SAT & CON \\
\midrule
Ratio of sentences containing adjectives & 0.75 & 0.70 & 0.52 & 0.56 & 0.71 \\ 
Ratio of sentences containing adverbs & 0.70 & 0.70 & 0.63 & 0.75 & 0.66 \\ 
Ratio of sentences containing conjunctions & 0.66 & 0.63 & 0.57 & 0.57 & 0.60 \\ 
Ratio of sentences containing numerals & 0.44 & 0.20 & 0.22 & 0.18 & 0.26 \\
Ratio of indefinite pronoun per sentence & 0.50 & 0.59 & 0.44 & 0.53 & 0.50 \\
Ratio of personal pronoun per sentence & 0.24 & 0.35 & 0.38 & 0.34 & 0.35 \\
%Pronoun ratio     & 0.49 & 0.53 & 0.50 & 0.54 & 0.52 \\
\hline
Lexical expressivity \cite{zhou2004comparison} & 0.32 & 0.40 & 0.32 & 0.34 & 0.38 \\
Ratio of modifiers (adapted from  \cite{zhou2004comparison}; norm. by \#content words) & 0.23 & 0.27 & 0.23 & 0.24 & 0.26 \\
Pausality \cite{zhou2004comparison} & 4.87 & 3.58 & 3.45 & 3.56 & 3.55 \\
Redundancy (adapted from \cite{zhou2020survey}; \#function words norm. by \#w) & 0.31 & 0.33 & 0.30 & 0.32 & 0.32 \\
%1S                       & 0.04 & 0.07 & 0.13 & 0.13 & 0.04 \\
%1P                       & 0.09 & 0.14 & 0.06 & 0.08 & 0.10 \\
Modality* (adapted from \cite{zhou2020survey}; norm. by \#s) & 0.11 & 0.13 & 0.07 & 0.13 & 0.12 \\
\hline %div by the number of sents
Ratio of headlines containing sentiment terms & 0.46 & 0.40 & 0.51 & 0.43 & 0.51 \\
%percentagem dos títulos que continham, ao menos, uma palavra de sentimento (SentiLex).
\bottomrule
\end{tabular}
\label{tab:data_stats_linguistic}
\end{threeparttable} 
\begin{tablenotes}
\item *Modality was estimated through the most frequent modal verbs in Portuguese (i.e., \textit{poder}, \textit{dever}, \textit{ter de}, \textit{precisar}), indicating (im)possibility, contingency, or necessity.
\end{tablenotes}
\vspace{-3mm}%Put here to reduce too much white space after your table 
\end{table*}

Table \ref{tab:most_common_terms_headlines} presents the top-10 most frequent content words in the headlines of each news category in MINT. In general, topics related to the COVID-19 pandemic permeated all categories, either directly or indirectly. Accordingly, the most frequent content words include \textit{covid-19} or \textit{covid}, or related terms, such as \textit{casos} (\textit{cases}), \textit{pandemia} (\textit{pandemic}), and \textit{vacina} (\textit{vaccine}), explicitly referring to the new coronavirus. In addition, it is interesting to stress the prominence of the terms \textit{novo} (\textit{new}) and \textit{ano} (\textit{year}) in almost all the categories, which are probably linked to the emergence of the \textit{new} virus and its impact in the \textit{year} considered in our corpus. Another characteristic shared by almost all categories is the use of terms related to the national geopolitical context, such as \textit{governo} (\textit{government}), \textit{país} (\textit{country}), portugueses (\textit{Portuguese}), and \textit{Portugal}. Interestingly, the satirical articles focus mainly on the political actors involved in the national agenda, such as the Portuguese President (\textit{Marcelo} [Rebelo de Sousa]), and the Prime Minister of Portugal ([António] \textit{Costa}). On the contrary, the soft news focus chiefly on popular reality shows (\textit{Big Brother}), entertainers (\textit{Cristina Ferreira}), and personal relationships through terms like \textit{filha} (\textit{daughter}), and \textit{filho} (\textit{son}). Moreover, this type of content explores sentiment and emotions, e.g., \textit{amor} (\textit{love}), and make use of predicates such as \textit{revela} (\textit{reveal}), which are usually found in clickbait titles. On opposite, the most frequent verb in hard news is the declarative form of the verb \textit{dizer} (\textit{say}), which is probably used to introduce citations in text. With regard to conspiracies, with the exception of the use of the qualitative adjective \textit{grande} (\textit{big}), and the reference to USA (\textit{EUA}), an important player in the global affairs, the most frequent terms are quite similar to the ones found in the hard news. This aspect is not surprising, since conspiracy approaches track news topics, and try to mimic real news.

\subsection{Insights from crowdsourced annotations}
\label{sec:human_annotation}

\begin{figure}[t]
%\vspace{-3mm}%Put here to reduce too much white space
    \centering
    \includegraphics[width=\textwidth]{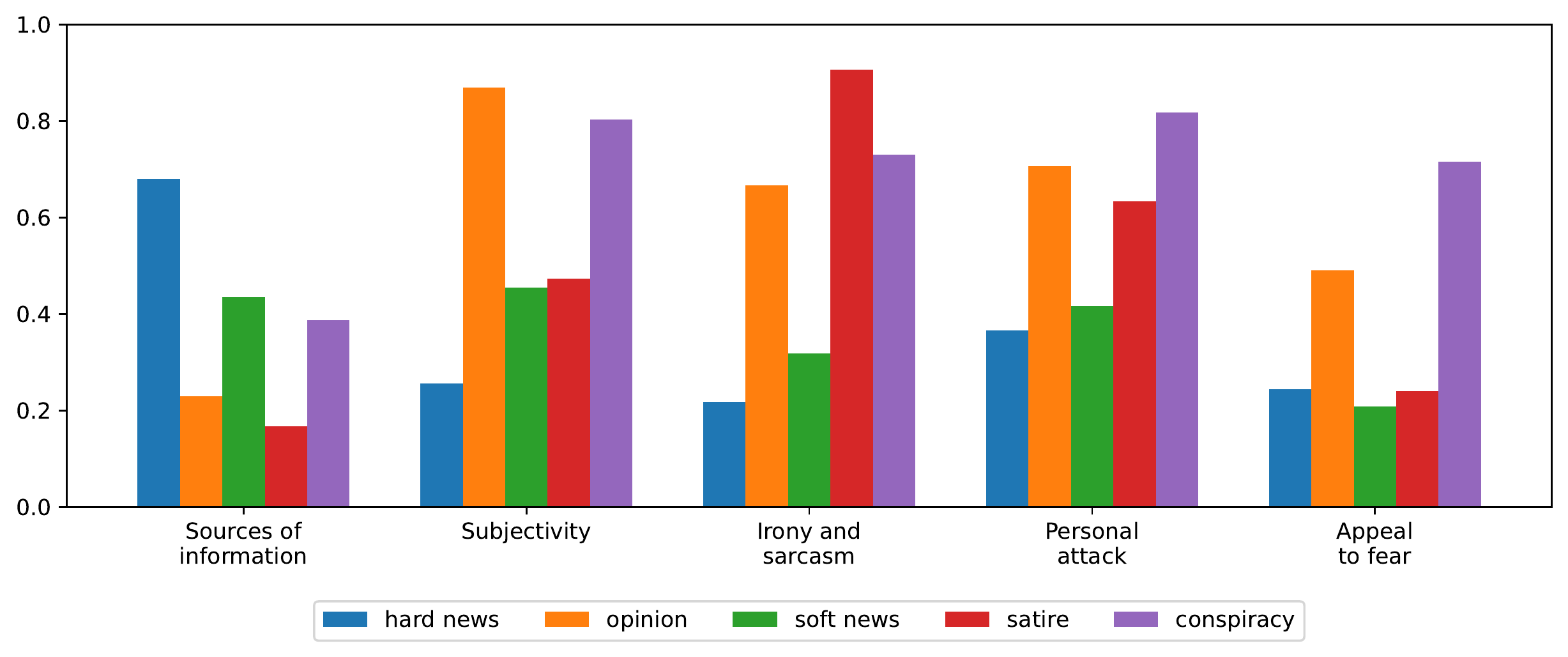}
    \caption{Relative frequency to the dichotomous questions for each MINT category.}
    \label{fig:dichotomous_questions}
\vspace{-3mm}%Put here to reduce too much white space after your table 
\end{figure}

Figure \ref{fig:dichotomous_questions} summarizes the answers to the dichotomous questions under the perspective of online news readers. The result reinforces the similarity between the opinion and conspiracy articles, also observed in Table \ref{tab:data_stats_linguistic}. The incidence of subjective information, a feature usually observed in opinion articles, also appears as a strong characteristic of conspiracies. Moreover, both categories present a high level of irony and/or sarcasm, and often use fallacies, in particular personal attack (i.e., the author attacks a specific individual or organization rather than attacking the substance of the argument itself). On the other hand, hard news usually follow the journalistic standards and practices, including accuracy (materialized, for instance, by the use of reliable sources of information), objectivity, and impartiality. Those characteristics are also observed in soft news, although to a lesser extent. As expected, users are capable of easily identifying irony and sarcasm in satirical news articles; however, their annotations also demonstrate that this property can be observed in multiple categories, namely in opinion news articles and conspiracy theories, as previously mentioned. Furthermore, the fallacious arguments typically used in conspiracy (namely, personal attacks and appeals to fear) can also be found in satirical and opinion articles. 

\begin{table*}[t]
\caption{10 most frequent content words in the headlines of each category.}
%\resizebox{\linewidth}{!}{
\label{tab:most_common_terms_headlines}
\begin{tabular}{l|l}
\toprule
H-N	& covid-19, portugal, governo, diz, casos, vai, novo, contra, anos, ser \\
OPI	& portugal, pandemia, futuro, ser, novo, país, covid-19, europa, governo, política \\
S-N & amor, big, brother, cristina, covid-19, filha, ferreira, filho, revela, anos \\
SAT & vai, portugueses, marcelo, ser, portugal, costa, novo, governo, vão, pessoas \\
CON	& covid-19, contra, pandemia, vacina, eua, covid, parte, portugal, vacinas, grande \\
\bottomrule
\end{tabular}
\vspace{-3mm}%Put here to reduce too much white space after your table 
\end{table*}

\vspace{-2mm}%Put here to reduce too much white space

\section{Conclusion and Future Work}
\label{sec:conclusion}

MINT, a corpus comprising news articles published by different Portuguese mainstream and independent sources, fills a gap in misinformation literature, providing annotated resources to enable studies ranging from social sciences to computational journalism. In particular, this corpus can help answering research questions involving the study of news credibility, and support the development of several NLP tasks, including automatic identification of misinformation, authorship attribution, and automatic detection of fallacies (for example, based on the conspiracy theories that surround the new coronavirus pandemic).
A forthcoming release of MINT will add other news categories, new sources, and include more annotated articles. 

%\newpage